\documentclass[journal]{IEEEtran}
\usepackage[font=small]{caption}

\usepackage{cite}

\ifCLASSINFOpdf
\usepackage[pdftex]{graphicx}
\else
\fi
\usepackage{caption}
\begin{document}
%
\title{Modeling Collapse of Steered Vine \\Robots Under Their Own Weight}
%
%
%
\author{Ciera~McFarland,~\IEEEmembership{Student Member,~IEEE}
        and~Margaret~McGuinness,~\IEEEmembership{Member,~IEEE}
\thanks{C. McFarland and M. McGuinness are with the Department
of Aerospace and Mechanical Engineering, University of Notre Dame, Notre Dame
IN, 46556 USA e-mail: {\tt\small \{cmcfarl2, mmcguinness\}@nd.edu}}
\thanks{
DISTRIBUTION STATEMENT A. Approved for public release. Distribution is unlimited.
This material is based upon work supported by the Department of the Air Force under Air Force Contract No. FA8702-15-D-0001 or FA8702-25-D-B002. Any opinions, findings, conclusions or recommendations expressed in this material are those of the author(s) and do not necessarily reflect the views of the Department of the Air Force. © 2025 Massachusetts Institute of Technology. Delivered to the U.S. Government with Unlimited Rights, as defined in DFARS Part 252.227-7013 or 7014 (Feb 2014). Notwithstanding any copyright notice, U.S. Government rights in this work are defined by DFARS 252.227-7013 or DFARS 252.227-7014 as detailed above. Use of this work other than as specifically authorized by the U.S. Government may violate any copyrights that exist in this work.}
}

%
%

\markboth{Journal of \LaTeX\ Class Files,~Vol.~14, No.~8, August~2015}%
{Shell \MakeLowercase{\textit{et al.}}: Bare Demo of IEEEtran.cls for IEEE Journals}
%



\maketitle

\begin{abstract}
Soft, vine-inspired growing robots that move by eversion are highly mobile in confined environments, but, when faced with gaps in the environment, they may collapse under their own weight while navigating a desired path. In this work, we present a comprehensive collapse model that can predict the collapse length of steered robots in any shape using true shape information and tail tension. We validate this model by collapsing several unsteered robots without true shape information. The model accurately predicts the trends of those experiments. We then attempt to collapse a robot steered with a single actuator at different orientations. Our models accurately predict collapse when it occurs. Finally, we demonstrate how this could be used in the field by having a robot attempt a gap-crossing task with and without inflating its actuators. The robot needs its actuators inflated to cross the gap without collapsing, which our model supports. Our model has been specifically tested on straight and series pouch motor-actuated robots made of non-stretchable material, but it could be applied to other robot variations. This work enables us to model the robot's collapse behavior in any open environment and understand the parameters it needs to succeed in 3D navigation tasks.   

\end{abstract}


%
\IEEEpeerreviewmaketitle

\section{Introduction}
%
%
%
%
\IEEEPARstart{R}{obots} can move through dangerous environments without the safety concerns that humans have. Soft robots are particularly suitable for this purpose because they can use their adaptable bodies to navigate complex spaces that rigid robots cannot fit into. Everting vine robots \cite{blumenschein2020design} are useful in cluttered environments because they lengthen from their tip, i.e., grow by eversion, without moving relative to the ground. In dangerous situations such as urban search and rescue, this type of motion can prevent the movement of unstable debris. These robots can squeeze through tight holes~\cite{hawkes2017soft}, function well in rubble \cite{der2021roboa}, \cite{mishima2006development}, and grow vertically in confined spaces \cite{CoadRAM2020}. These robots steer using a variety of actuators, such as tape \cite{hawkes2017soft}, latches \cite{hawkes2017soft}, magnets \cite{li2021bioinspired}, strings that get cut \cite{cinquemani2020design}, tendons \cite{blumenschein2018helical}, internal steering devices \cite{der2021roboa}, and pneumatic actuators \cite{hawkes2016design,greer2017series,naclerio2020simple,CoadRAM2020,kubler2023multi}. However, while this gives them high functionality in two-dimensional or cluttered spaces, they currently struggle to reliably navigate open three-dimensional environments because of their limited ability to support loads against gravity, including their own weight.  

\begin{figure}[tb]
\centering
\includegraphics[width = \columnwidth]{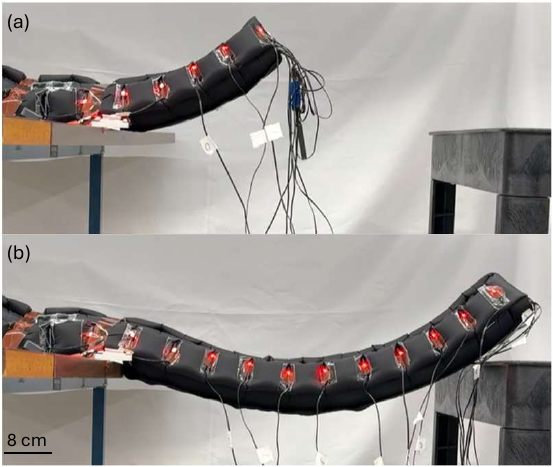}
\caption{Demonstration where our steered vine robot collapse model successfully predicts a lack of collapse during a gap-crossing task. (a) The robot curves to avoid deflecting off course. (b) The curve enables the robot to grow high enough such that when it deflects, it lands on the other side of the gap instead of deflecting underneath it or collapsing.}
\label{gs}
\vspace{-3ex}
\end{figure}

In open spaces, vine robots are susceptible to collapse, either at their base due to a transverse load or at their center due to an axial load. There are existing models to predict transverse collapse \cite{leonard1960structural}, \cite{comer1963deflections} and axial buckling \cite{fichter1966theory} of inflated beams due to external weights. These models have previously been validated for vine robots loaded with external weights \cite{luong2019eversion,hwee2023kinematic}, undergoing retraction \cite{coad2020retraction}, and contacting their environments \cite{greer2018obstacle,greer2020robust,haggerty2019characterizing}. One study, \cite{mcfarland2023collapse}, is to our knowledge the only work to consider whether these collapse models can be applied to an unsteered vine robot loaded only with its own weight, which was generally considered negligible compared to other external forces. In that study, the authors found that unsteered vine robots can collapse under their own weight, and they will collapse at longer lengths when they are grown at steeper growth angles, higher internal body pressures, or larger body diameters.

However, an analysis of strictly unsteered robots is insufficient for operating vine robots in the real world. Not every task can be achieved using straight-line motion. Whether this is because the robot is actively steering \cite{hawkes2017soft},\cite{greer2019soft}, turning due to obstacle interactions \cite{greer2018obstacle}, \cite{greer2020robust}, or simply deflecting as observed in \cite{mcfarland2023collapse}, the robot will eventually curve. An example of this is seen in Figure~\ref{gs}. Here, the robot must steer to reach a target that requires crossing a gap. Without steering, it could deflect and fail to reach the target. Curving upward ensures that it is higher than the ledge as it grows (Figure~\ref{gs}(a)), and when it falls significantly due to deflection, it has already cleared the gap (Figure~\ref{gs}(b)). We need a model that can predict when a robot in a steered shape will collapse under its own weight so that we can ensure that a robot's design will allow success in a given task. 

In some designs, helium has been used to prevent collapse of a robot turning and moving through free space \cite{takeichi2017development}, \cite{kaleel2023framework}, but this may not be a viable option in the types of situations where we envision the robot being deployed due to limited resources and time to respond. Vine robots can also be stiffened through the use of pneumatic pouches \cite{do2024stiffness},\cite{feteih2025active},\cite{wang2022geometric} or other stiffening substances \cite{fuentes2023deployable}. This does increase their ability to resist collapse, but also makes the robot heavier. None of these works model a robot's collapse length under stiffened conditions, so our current work can serve as the first step towards determining whether adding stiffness is a viable solution to prevent for vine robots collapsing under their own weight in free space.

This work builds upon the initial study in \cite{mcfarland2023collapse}, which analyzed the collapse of unsteered vine robots under their own weight. Here, we contribute a general, shape-aware collapse framework applicable to both steered and unsteered configurations. Specifically, we:

\begin{itemize}
    \item Develop a tail tension-aware model that generalizes the base moment formulation to account for the internal tail force. 
    \item Introduce a comprehensive collapse model that unifies the behavior of unsteered and steered vine robots and accurately predicts collapse across a wide range of parameters and geometries.
    \item Provide extensive new experimental validation on multiple robot designs, including steered robots, confirming the model’s predictive accuracy.
    \item Demonstrate the model’s field relevance through tasks that require steering, showing how the framework enables prediction and prevention of collapse in realistic environments.
\end{itemize}

Together, these advances establish a general predictive framework for vine-robot stability in arbitrary shapes, providing the foundation for model-based design and planning of future growing robot systems.

\section{Unsteered Robot Collapse} \label{previous work}
In this section, we present the model for the length at which an unsteered vine robot (modeled as a double-layer cylinder to represent its outer wall and inner tail) will collapse under its own weight given a certain growth angle, internal body pressure, and robot body diameter, when accounting for the robot's tail tension. In this study, we are only considering collapse at the base due to transverse loading conditions, as axial loading conditions are unlikely to occur in this context. We build on the previous model presented in \cite{mcfarland2023collapse} by adding tail tension, an additional force caused by the robot's inner tail pulling backwards against the direction of growth \cite{coad2020retraction}. We also extend the range of parameters tested to highlight the accuracy of this new model under a variety of conditions. This includes broadening the range of pressures and diameters tested, as well as testing more combinations of angles, pressures, and diameters.

\subsection{Model Without Tail Tension} \label{straight model}
\begin{figure}[tb]
\centering
\includegraphics[width = \columnwidth]{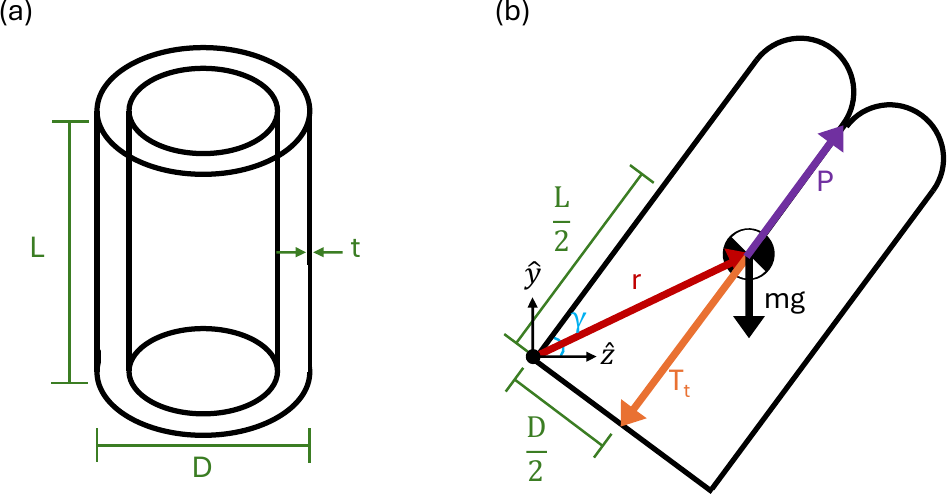}
\caption{Diagram of vine robot parameters used in our collapse model (modified from \cite{mcfarland2023collapse}). (a) The robot consists of two layers of material with length $L$, diameter $D$, and thickness $t$. (b) The robot, growing at an angle $\gamma$ above the horizontal, has three forces acting on it in the directions shown: gravity pointing down ($mg$), tail tension pointing back toward the base ($T_t$), and air pressure ($P$) pointing in the direction of growth. Air pressure acts on all walls of the robot, but the net force points toward the tip. The center of mass is a distance $r$ from the base of the robot.}
\label{moment diagram}
\vspace{-3ex}
\end{figure}

Leonard~\cite{leonard1960structural} and Comer and Levy~\cite{comer1963deflections} define the moment needed to transversely collapse an inflated beam as 

\begin{equation} \label{eq:straight collapse moment}
M_{collapse} = \frac{P\pi D^3}{8},
\end{equation} where \(P\) is the internal pressure and \(D\) is the inflated diameter. In Figure~\ref{moment diagram}, we show the parameters of the vine robot when viewed as a two-layer inflated tube with three forces acting on it: gravity, tail tension, and internal pressure. If we sum the moments from these forces, we should be able to determine the length at which a vine robot will collapse.

Gravity acts at the center of mass of the robot, creating a positive moment about the $x$-axis at the pivot point shown in Figure~\ref{moment diagram}. This moment is described by

\begin{equation} \label{eq:Mweight}
M_{weight} = mgr_z,
\end{equation} where \(m\) is the robot's mass, \(g\) is the gravitational constant, and \(r_z\) is the projection of the position vector from the point where the robot will fold during collapse to its center of mass onto the $z$-axis. The position vector, \(r_z\), can be defined by

\begin{equation} \label{eq:r vector}
r_z = \frac{D}{2}\sin{\gamma}+\frac{L}{2}\cos{\gamma},
\end{equation} where \(\gamma\) is the growth angle, and \(L\) is the length of the robot.

 To determine the robot's mass, we first multiply the modeled cylinder's volume by its material density. We double this quantity to account for there being two layers of fabric in each segment.  This is defined by \begin{equation} \label{eq:robot mass}
m = 2\pi DtL\rho,
\end{equation} where \(t\) and \(\rho\) are the thickness and the density of the body material, respectively.

At the instant of collapse, the moment due to weight counteracts the other forces acting on the robot to prevent collapse. Wrinkles propagate around the robot's circumference, causing it to fold at its topmost point like a hinge. Therefore, we assume the robot is static and use Newton's second law to set Equation~\ref{eq:Mweight} equal to Equation~\ref{eq:straight collapse moment}. Then, we plug Equations~\ref{eq:r vector} and \ref{eq:robot mass} into their respective places in Equation~\ref{eq:Mweight}. This makes it possible to solve for the length at which a two-layer robot will collapse given a certain growth angle, robot body diameter, internal body pressure, body material thickness, body material density, and gravitational constant as

\begin{equation} \label{eq:full straight model}
L = \frac{-D}{2}\tan{\gamma} + \frac{\sqrt{\pi D^2\rho gt\sin^2{\gamma}+\frac{1}{2}P\pi D^2\cos{\gamma}}}{2\cos{\gamma}\sqrt{\pi\rho g t}}.
\end{equation}

\subsection{Tail Tension Adjusted Model}

\begin{figure}[tb]
\centering
\includegraphics[width = 0.75\columnwidth]{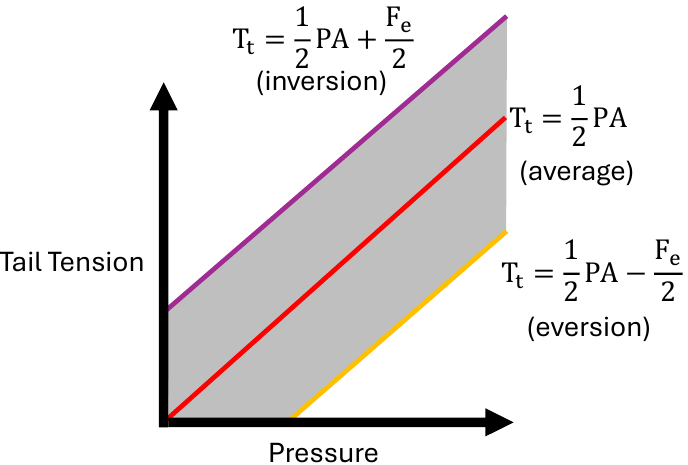}
\caption{Plot of expected tail tension values varying with pressure for a quasistatic vine robot as used in our experiments. For a quasistatic robot, tail tension is approximately 
half of the robot's internal body pressure \(P\), multiplied by its area \(A\). However, uncertainty about how the pressure force is shared between the robot's tail and wall can cause the tail tension to be higher or lower than that average value. The maximum amount by which a quasistatic robot will deviate from that value is\(\frac{F_e}{2}\), that is, the force needed to evert the robot. The purple curve shown represents the point at which the robot begins inverting. The yellow curve represents the point at which the robot begins everting or growing, and we expect the tail tension value for our robot to be approximately along this line. In between the two boundary curves, the robot is stopped. The red curve represents the average value of tail tension, which is between the two extremes. It is possible to achieve values outside the boundary curves by inverting or everting quickly.}
\label{tail tension diagram}
\vspace{-3ex}
\end{figure}

Tail tension has a variable range of values where it can be~\cite{coad2020retraction}. Equation~\ref{eq:tail tension} describes the expected range of values for tail tension if it is everting as slowly as possible, stopped, or inverting as slowly as possible. If the tail tension value is at the minimum value in the range, the robot is everting. If it is at the maximum value in that range, the robot is inverting. At any other value in the range, it is stopped. The range is meant to give us reasonable expectations for a vine robot's variable behavior in a quasistatic state. This range is shown in Figure~\ref{tail tension diagram}. It is possible to achieve values outside of the shown range by inverting or everting quickly, but we do not consider such dynamic motions in this paper.

The value of tail tension for the quasistatic case is approximately 

\begin{equation} \label{eq:tail tension}
\frac{1}{2}\frac{P\pi D^2}{4} - \frac{F_e}{2} \leq T_t \leq \frac{1}{2}\frac{P\pi D^2}{4} + \frac{F_e}{2},
\end{equation} where $\frac{1}{2}\frac{P\pi D^2}{4}$ is the average value of tail tension (assuming the force due to the pressure at the tip of the robot is counteracted equally by tension in the tail and the wall), and $F_e$ is a correction factor that represents the minimum force required to begin everting the robot under any conditions, including the addition of inflated actuators or when the robot is otherwise not straight. This term includes both curvature and length-dependent factors that can also be modeled separately~\cite{blumenschein2017modeling}. Our term $F_e$ is defined by

\begin{equation} \label{eq:Fe}
F_e = \frac{P_e\pi D^2}{4},
\end{equation}

where $P_e$ is the minimum pressure for the robot to begin growing. This term, $F_e$ is similar but slightly different from the term $2 F_i$ used in \cite{coad2020retraction}, which was the force to evert a short, unsteered robot. That force to evert the robot is proportional to the material thickness squared \cite{girerd2024material}. For our $F_e$ term, we experimentally determined it for every experiment, regardless of the conditions of the robot. This will be explained in more detail in Section~\ref{Fe section}.

Tail tension is expected to act along the robot's central axis. Therefore, we assume that this force is applied at a distance $\frac{D}{2}$ from the robot's collapse point, and we subtract it from the previous $M_{collapse}$, which gives us a new value of 

\begin{equation} \label{eq:tail tension moment}
M_{collapse} = \frac{1}{2}\frac{P\pi D^3}{8}\pm \frac{F_eD}{4}.
\end{equation} This then modifies our predicted collapse length to be 

\begin{equation} \label{eq:full straight model with tail tension}
L = \frac{-D}{2}\tan{\gamma} + \frac{\sqrt{\pi D^2 \rho g t\sin^2{\gamma} + (\frac{1}{4}P\pi D^2 \pm F_e)\cos{\gamma}}}{2\cos{\gamma} \sqrt{\pi \rho gt}}. \end{equation} 

\subsection{Robot Fabrication} \label{fabrication straight}
In this section, we fabricated all robots from 40 denier TPU-coated heat-sealable ripstop-nylon (extremtextil, Dresden, Germany). We used calipers to measure the material thickness as 0.031~mm and weighed the fabric to calculate a density of 2200~kg/m\(^3\). We cut each robot from a flat bolt of fabric, formed it into a tube, and sealed it along its length using an impulse sealer (Jorestech, Sunrise, FL) with a 0.5 cm seal width. We sealed each robot at both ends and then sealed it again such that the inner tail was sealed to the bottom of the outer wall at the robot base to prevent eversion. We excluded this step for robots that needed to grow, such as the short robots described in Section~\ref{Fe section}. We cut each robot 2.54~cm wider than necessary to ensure it was the appropriate diameter. This left an extra flap of material approximately 1.27~cm long on either side of the long seam. The weight of this flap is added to the model by increasing mass to be

\begin{equation}
\label{eq:robot mass plus fin}
m = 2(\pi D +f) tL \rho,
\end{equation} where, for a given cross-section, \(f\) is twice the height of the robot flap, which is approximately 0.03~m based on the description above. The height is doubled to account for there being two flaps. This modified version of Equation~\ref{eq:robot mass} is then incorporated into Equations~\ref{eq:full straight model} and \ref{eq:full straight model with tail tension}. 

\subsection{Determination of Eversion Force} \label{Fe section}
For our unsteered robots, we found it to be about 1.4~N when the robot has the flap of extra material around its seam discussed in \cite{mcfarland2023collapse}. We determined this value by building short (less than 0.25~m) robots of the same material as the robots used in these experiments and inflated diameters of 2.43~cm, 3.24~cm, 4.04~cm, and 4.85~cm, as these were the initial robot diameters tested in \cite{mcfarland2023collapse}. These robots also had the flap at their seam to imitate the full-size robots used in these trials. Starting from 0~kPa, we increased the pressure in these robots in increments of 0.35~kPa until the robots began to grow consistently. We repeated these trials several times for each diameter. We then plotted these pressure-to-grow values against the inverse of their corresponding robot area. The slope of this curve is \(F_e\), the force to evert the robot. 

\subsection{Experimental Setup}
\begin{figure}[tb]
\centerline{\includegraphics[width = \columnwidth]{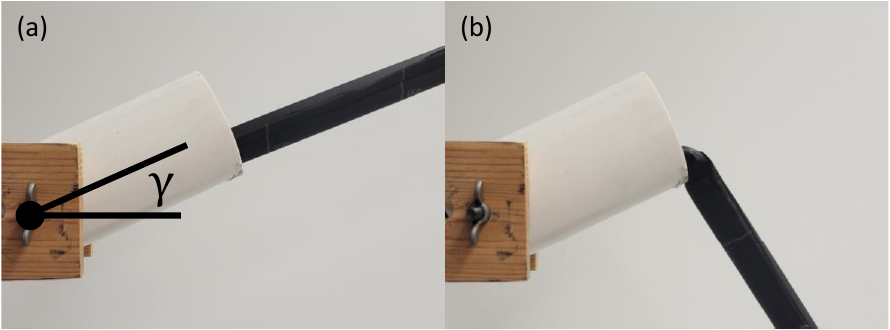}}
\caption{Experimental setup for positioning vine robots at various angles, used to obtain the results shown in Figure~\ref{giant summary plot} (reprinted from \cite{mcfarland2023collapse}). (a) The robot is pushed out of the pipe, which is positioned at some adjustable growth angle \(\gamma\) above horizontal. (b) If the robot collapses due to transverse loading, it will do so at the edge of the pipe, which is the last point of support.}
\label{angle pipe}
\vspace{-3ex}
\end{figure}

For the following experiments, we aimed to reanalyze the work in \cite{mcfarland2023collapse}, where they tested the collapse length of vine robots at different growth angles, pressures, and diameters. For the growth angle trials in \cite{mcfarland2023collapse}, the authors lengthened the robot by pushing it at a consistent angle through an angled pipe until it collapsed. This angled pipe, shown in Figure~\ref{angle pipe}, served as the last point of support for the robot, which is the location where collapse should occur. They did not test negative angles steeper than \mbox{-65$^{\circ}$} because the robot tended to deflect and not actually collapse at those trajectories. That robot had a diameter of 2.43~cm and an internal pressure of 3.45~kPa. In  Figure~\ref{giant summary plot}(a), we present the same data from \cite{mcfarland2023collapse}, excluding the axial collapse data, which occurs at an angle of \mbox{90$^{\circ}$}, but reanalyzed through the lens of our models. The axial case was excluded as it is unlikely to occur.   

Next, we measured the collapse length of a vine robot with a fixed growth angle and diameter, but varying pressure. In \cite{mcfarland2023collapse}, the robot was set to a specific length, and the pressure was either increased or decreased until the robot collapsed or uncollapsed. In these trials, we used the same procedure as the above trials, meaning the robot is lengthened by pushing until collapse occurs. The first group of trials had the same parameters as the one in \cite{mcfarland2023collapse}, where the robot had a growth angle of \mbox{0$^{\circ}$} and a diameter of 2.43~cm. However, they only evaluated pressures up to a maximum of 13.8~kPa. Here, we extended the range to be from 2.1~kPa to 27.8~kPa. The second group of trials had a robot with a growth angle of \mbox{45$^{\circ}$} and a diameter of 2.43~cm. These results are shown in Figure~\ref{giant summary plot}(b) and (c).

Finally, we measured the collapse length of vine robots with fixed growth angles and internal pressures, but varying diameters. In \cite{mcfarland2023collapse}, they tested inflated diameters of 2.43~cm, 3.24~cm, 4.04~cm, and 4.85~cm at a growth angle of \mbox{0$^{\circ}$} and an internal pressure of 2.07~kPa. Here, we present new trials from that setup, along with two additional diameters, 1.21~cm and 8.49~cm. Since we were extending the range of parameters being evaluated, we opted to take an entirely new set of data as opposed to presenting the data from \cite{mcfarland2023collapse}. As the growth angle was \mbox{0$^{\circ}$}, we pushed the robot off the table instead of using the launching pipe. We also performed this experiment with a growth angle of \mbox{0$^{\circ}$} and an internal pressure of 6.89~kPa, and a growth angle of \mbox{45$^{\circ}$} and an internal pressure of 2.07~kPa. This last round of experiments again featured the use of the launching pipe. Robots with a diameter greater than the launching pipe's were pushed along the top of the launching pipe. The results of these trials are shown in Figure~\ref{giant summary plot}(d)-(f).

\subsection{Results}\label{Unsteered Results}

\begin{figure*}[tb]
\centering
\includegraphics[width=\textwidth]{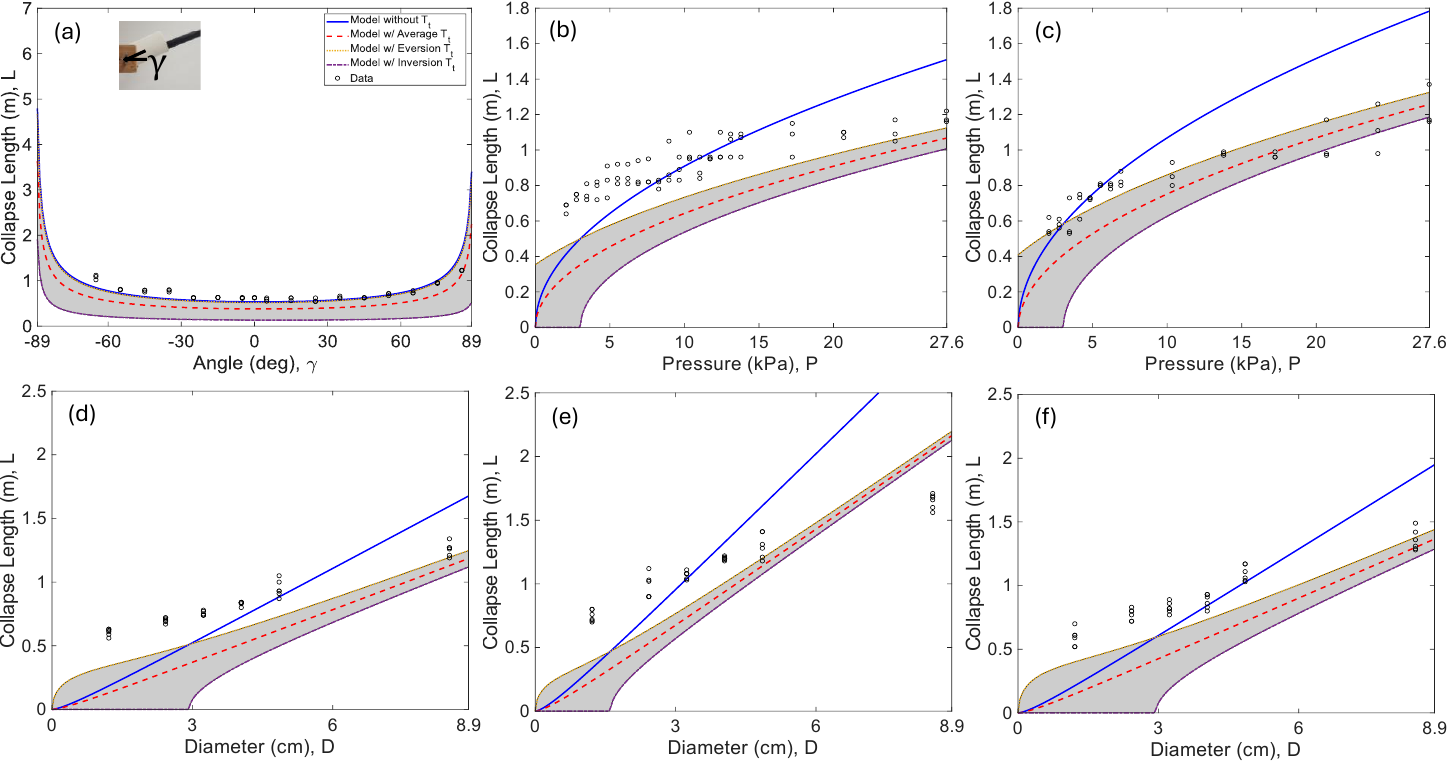}
\caption{Results of how collapse length varies with growth angle, pressure, and diameter for unsteered fabric robots. (a) Collapse length vs. growth angle plot for a 2.43~cm diameter robot at 3.45~kPa (modified from \cite{mcfarland2023collapse}). Collapse length vs. pressure plots for a 2.43~cm diameter robot at growth angles of (b) \mbox{0$^{\circ}$} and (c) \mbox{45$^{\circ}$}. Collapse length vs. diameter plots for (d) a growth angle of \mbox{0$^{\circ}$} and a pressure of 2.07~kPa, (e) a growth angle of \mbox{0$^{\circ}$} and a pressure of 6.89~kPa, and (f) a growth angle of \mbox{45$^{\circ}$} and a pressure of 2.07~kPa. The predicted collapse lengths for our model incorporating no tail tension (blue solid line), average tail tension (red dashed line), eversion tail tension (yellow dotted line), and inversion tail tension (purple dash dot line) are shown along with the data (black circles). The eversion tail tension model has the best match with all six configurations. The latter three models mentioned above are also shown in Figure~\ref{supports plot}.}
\label{giant summary plot}
\vspace{-3ex}
\end{figure*}

In all trials, we present four models of collapse: no tail tension (blue solid line), average tail tension (red dashed line), eversion tail tension (yellow dotted line), and inversion tail tension (purple dash dot line). If the collapse length predicted by the inversion tail tension model was negative, this value was adjusted to zero on the plots. This indicates that pulling backwards on the tail at any point during that trial would have caused the robot to collapse, regardless of length.

In Figures~\ref{giant summary plot}(a)-(f), we see that the eversion tail tension curve has the best match of the data's slope, though it is downshifted from the data in most of these plots. In Figure~\ref{giant summary plot}(a), the robot does not have sufficient pressure to grow and the model without tail tension is also very accurate. In Figures~\ref{giant summary plot}(b) and (c), the robot would have sufficient pressure to grow around 4~kPa. Prior to that, we see that the no tail tension curve sometimes has better agreement with the data than the eversion tail tension curve. In Figures~\ref{giant summary plot}(d) and (f), all but the two smallest robots have sufficient pressure to grow. In Figure~\ref{giant summary plot}(e), all robots are able to grow. From this, it is reasonable to assume that the eversion tail tension model should have the best fit.

Despite this, the data often overshoots the model. This is likely due to excessive deflection under gravity, which the model assuming the robot takes on a cylindrical shape does not account for. For robots growing at an angle near the horizontal, deflection would make them have a shorter moment arm, allowing them to achieve a longer length before eventually collapsing. This could explain why the eversion tail tension model largely describes the correct trend, but appears to be downshifted from the data in several cases where the angle is close to the horizontal. Deflection could also explain why the largest diameter in Figure~\ref{giant summary plot}(e) does not appear to be on the same trajectory as the other diameters. In this configuration, all models fail to slope match the data, but the largest diameter still appears to have collapsed at a shorter length than would be expected from looking at the other diameters. It has the longest collapse length of any diameter trials and is the heaviest robot, so it could reasonably be expected to have the most deflection.

It is important to note that extending the pressure range was a key adjustment for understanding the impact of pressure on collapse length, as the model without tail tension matches the data in Figure~\ref{giant summary plot}(b) reasonably well in the lower half of the range, but overshoots it in the second half. In \cite{mcfarland2023collapse}, which only had the lower half of the range, it was believed the model was reasonable for the data. The overshoot in the upper half makes it clear this was not true and a new model was needed. However, the persistent overshoot of the data, even with the new model, indicates that some way to account for the true shape of the robot would likely improve accuracy.

\section{Unsteered Robot with Inflated Support Collapse} \label{inflated support section}
In this section, we analyze how an unsteered vine robot with inflated support structures collapses under its own weight. Inflated actuators are pressurized, which means they stiffen the robot. However, they also create wrinkles to turn the robot, which could weaken it. However, these are the actuators we ultimately want to use to steer the robot, and we will analyze how well the model predicts collapse in these robots in Section~\ref{steered section}. As a preliminary step before that, to separate the effects of pressurizing and turning, we devised an experiment to determine if our model can predict the collapse length of unsteered robots that have been stiffened through the use of additional pressurized supports, but are not turning.

\subsection{Modeling}
\begin{figure}[tb]
\centering
\includegraphics[width=0.75\columnwidth]{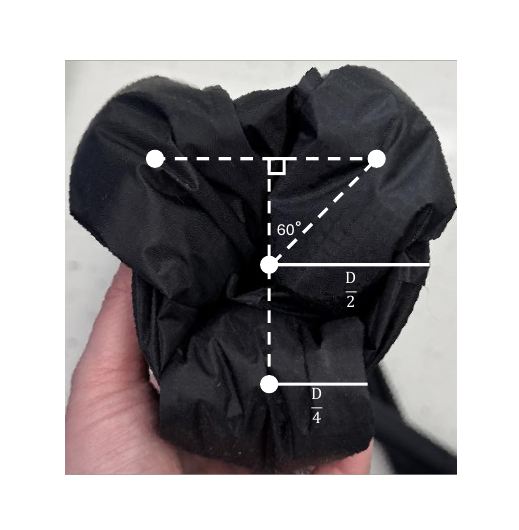}
\caption{Cross-sectional diagram of a vine robot with inflated tube supports, used to obtain the results shown in Figure~\ref{supports plot}. The three inflated tube supports are half the diameter of the robot and circular in cross-section; they are not sealed into series pouch motor actuators, as in Figures~\ref{gs}, \ref{actuated figure}, and \ref{New Demo}. One is located at the bottom of the robot with the other two being \mbox{120$^{\circ}$} apart from it in each direction around the robot's circumference.}
\label{inflated tube diagram}
\vspace{-3ex}
\end{figure}

In order to mimic the pressure and weight-based effects of adding steering actuators onto the robot body, we model our inflated support structures as if they are the same in quantity and location as the steering actuators used in previous work \cite{CoadRAM2020},\cite{mcfarland2024field}. That is, three actuators are \mbox{120$^{\circ}$} apart with one being aligned with the bottom of the robot. This actuator arrangement makes it possible to steer the robot in any direction with two degrees of freedom. Having two on the top helps counteract the robot's weight when lifting the tip. A diagram of this cross-section can be seen in Figure~\ref{inflated tube diagram}. 

We model each inflated tube support as having a circular cross-section, because it has not been heat-sealed into an actuator, i.e., it has not been compressed into a smaller cross-sectional area. We model the mass of this robot with inflated tube supports by modifying Equation~\ref{eq:robot mass} to be 

\begin{equation} \label{eq:robot and actuator mass}
m = 2 \left( \pi D+\frac{3\pi D}{2}\right)tL\rho + (0.044~\mathrm{m^{-1}})L, 
\end{equation} because there are three actuators that are half the diameter of the robot's body. In addition, tape is used to attach the actuators to the robot's body. Each of the three strips of tape weighs approximately 0.007~kg/m and is on the inside and outside of the robot, which accounts for the constant in Equation~\ref{eq:robot and actuator mass}. The previously mentioned extra flaps of material around the seam have been removed for this robot to reduce the weight.

With the mass adjusted, we then need to add in the effects of the inflated tube support pressure on the collapse moment. Each inflated support has a contribution of

\begin{equation} \label{support moment}
M_{support} = P_sA_sR_s, 
\end{equation} where \(P_s\) is the support's pressure, \(A_s\) is the support's area, and \(R_s\) is the distance from the support's center of mass to the point of collapse, the top of the robot body. Given the distances shown in Figure~\ref{inflated tube diagram}, the total contribution of these three inflated supports is \(\frac{3P_s\pi D^3}{32}\) assuming all the supports have the same pressure and area. Including tail tension, the moment to collapse the robot is now 

\begin{equation} \label{eq:inflated support collapse moment}
M_{collapse} = \frac{1}{2}\frac{P\pi D^3}{8} + \frac{3P_s\pi D^3}{32}\pm \frac{F_eD}{4}.
\end{equation} 

\subsection{Experimental Setup}
To test the model, we built a robot with an inflated diameter of 8.49~cm and three inflated support tubes half the robot's diameter. We pushed the robot off a table at a growth angle of 0$^{\circ}$ and an internal body pressure of 3.45~kPa. We kept all three inflated supports at the same pressure as each other for each trial, and we conducted three groups of trials where the pressure in the inflated supports was 0~kPa, 1.38~kPa, and 2.76~kPa. In this way, we tested if the model is accurate when the inflated supports are not pressurized at all, as well as when they are at some low arbitrarily chosen pressures. We ran six trials of each configuration. 

The force to evert the robot, \(F_e\), should have a different value for every pressure the inflated support tubes can be. We experimentally found this value to be 7.9~N, 7.9~N, and 11.1~N, respectively, for our three configurations. The value being the same for 0~kPa and 1.38~kPa inflated support pressure is possible because the pressure is low enough that the tubes do not impede the robot's ability to grow. For purposes of having a smooth model for the pressures we did not test $F_e$ at, we simply varied $F_e$ linearly between 8~N and 11~N for the pressure range of interest, which was 0~kPa to 3.45~kPa. The results of these trials can be seen in Figure~\ref{supports plot}.   

\subsection{Results}
\begin{figure}[tb]
\centering
\includegraphics[width = \columnwidth]{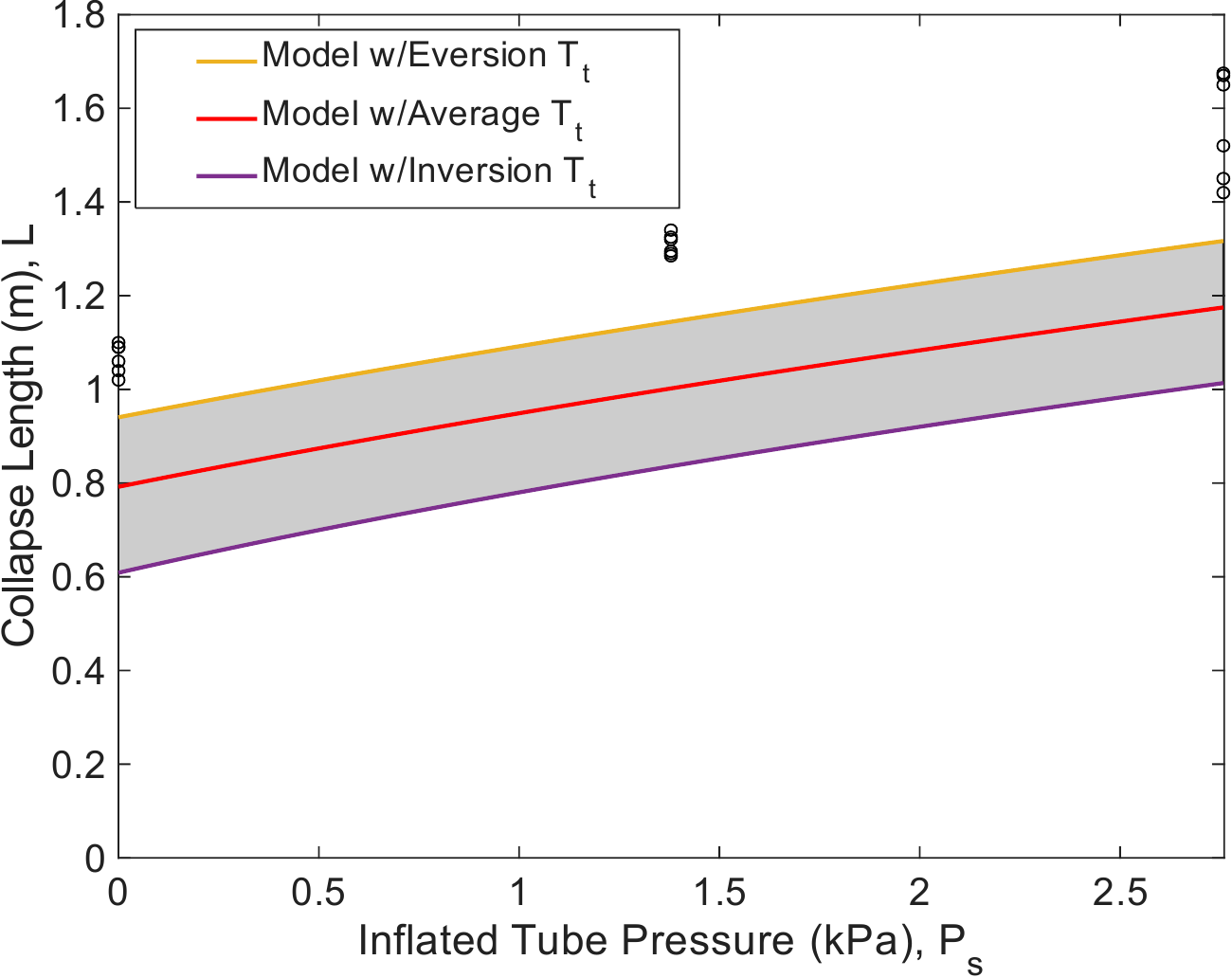}
\caption{Plot of collapse length vs. inflated tube supported pressure for an unsteered fabric robot with an 8.49~cm diameter and internal pressure of 3.45~kPa. Its supports varied in pressure from 0~kPa to 2.76~kPa. The eversion tail tension model matches the data best, which makes sense as we expect a robot of this size to grow at this internal pressure.}
\label{supports plot}
\vspace{-3ex}
\end{figure}

 In Figure~\ref{supports plot}, we show only the curves produced by the average tail tension curve (red dashed line), eversion tail tension model (yellow dotted line), and inversion tail tension curve (purple dash dot line). The model without tail tension was included in Figure~\ref{giant summary plot} to show how we have improved the analysis in this new work. From this point forward, we will be excluding it since we know it is inaccurate and there are no trials from \cite{mcfarland2023collapse} to compare these trials to. In general, the data matches the slope of all three curves, but all models are downshifted from the data. The eversion curve is the closest to the data, and we would expect this robot to grow given its interna pressure. Similar to the data shown in Figure~\ref{giant summary plot}, the gap between the model and the data could be due to deflection. This robot collapsed at lengths greater than 1~m and was significantly heavier than the robots in Section~\ref{Unsteered Results} due to its large diameter. This makes deflection highly likely, and that could explain why the data outperforms the model. Additionally, having inflated structures on the side of the robot does appear to prevent wrinkle formation, so it is possible these overperforming tests show that, under certain circumstances, the stiffening effect of the inflated supports is better than the model predicts. Since the data is still following the trends predicted by the model, we consider this sufficient evidence to support that the model can handle the effect of pressurized supports.

\section{Comprehensive Collapse}\label{comp collapse section overall}
In Section~\ref{previous work}, we established that adding tail tension to our unsteered robot model improves its accuracy. In Section~\ref{inflated support section}, we showed that more information can be added to this model such that it can apply to unsteered robots with pressurized supports. However, all of this work applies to robots that are assumed to be straight. In order to model the behavior of more complex vine robots, including ones that can steer, we need a comprehensive model that accounts for the robot's true shape. In this section, we present that model in its most general form, as well as how that model can be applied to unsteered and steered robots. 

\subsection{Comprehensive Collapse Model}
\label{comprehensive model section}
To begin examining the collapse behavior of robots with true shape information, we must first determine necessary changes for our model such that it can incorporate that information.
\subsubsection{Modeling}
\begin{figure}[tb]
\centering
\includegraphics[width=\columnwidth]{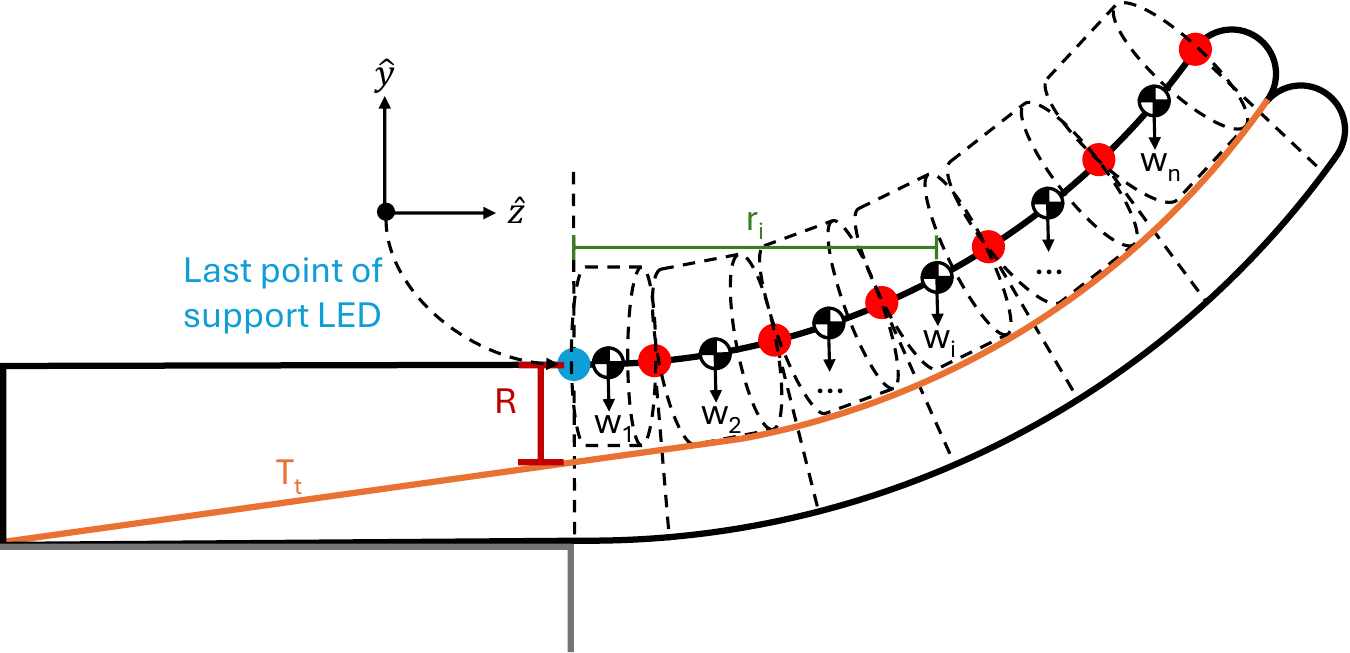}
\caption{Diagram of how current moment due to gravity in the positive $x$-direction about the base is calculated when the robot is modeled with discrete segments, as in all the following experiments. The blue motion capture LED marks the last point of support between the robot and its launching surface at the instant of collapse, and is the point from which all distances are measured. The segments of the robot are modeled as cylinders with a diameter $D$ equal to that of the robot, but with their centers in line with the red motion capture LEDs and their end points defined by the red motion capture LED locations. Each segment's weight, $w_i$, is modeled to be concentrated at the center of the segment, and the moment arm due to the weight of the segment about the base is the length $r_i$. As the robot's tail will usually take the shortest path from the robot's tip to its base, tail tension, $T_t$, is modeled to act in the negative $z$-direction at some distance $R$ from the blue LED, modeled as $\frac{D}{2}$.}
\label{segment moments}
\vspace{-3ex}
\end{figure}

Previously, we assumed that the robot was completely straight, which made it possible to model its distributed mass as a point mass located at the robot's midpoint. When considering a steered shape, this is no longer accurate. If the robot were steered into a constant curvature arc, we could theoretically model the distributed mass as being located at that arc's centroid. However, not all vine robots steer into constant curvature shapes, especially under the effect of gravity. We opted instead for a method that could be applied to any steered configuration. In this method, we view the vine robot as a series of discretized straight segments and sum up the moment acting on each segment about the robot's base. We track this shape using motion capture LEDs to divide up the segments. A diagram of these discretized segments aligned along the path of the motion capture LEDs is shown in Figure~\ref{segment moments}. Using this method, we can compare the current moment due to gravity acting on the robot to the moment that should collapse the robot according to our models. 

For the experiments described in Sections~\ref{straight with sensors section} and \ref{steered section}, we define the current moment due to gravity acting on the robot as the moment in the positive $x$-direction about the point on the top of the robot that is located at the last point of support before collapse. We have not observed the robot twisting into a state of collapse as a result of moments not in the $x$-direction when the robot is growing from a stationary base. 

In order to find the current moment due to gravity acting on the robot, we model each segment of the robot as a cylinder, just as we did in Sections~\ref{previous work} and \ref{inflated support section}. To solve for the current moment due to gravity from that particular segment, we multiply the doubled weight by the distance in the $z$-direction from the center of mass of the segment to the base of the robot where collapse will occur (blue LED in Figure~\ref{segment moments}) ($r_i$). The current moment due to gravity acting on the robot can be described by 

\begin{equation} \label{eq:current moment}
M_{current} = \sum_{i=1}^{n}2\pi(D+ND_{act})td_i\rho gr_i,
\end{equation} where $D$ is the inflated diameter of the robot, $N$ is the number of actuators, $D_{act}$ is the inflated diameter of the actuators, assuming they are circular and identical, $t$ is the material thickness, \(d_i\) is the length of a given segment, $\rho$ is the material density, and $g$ is the gravitational constant. 

The moment needed to collapse the robot is described by a more general version of Equation~\ref{eq:tail tension moment}. This moment has contributions from the pressure acting on the robot, the tail tension, and the actuators if present. Previously, since the top of the robot's body was the collapse point, the contributions of the robot's internal pressure and the tail tension were assumed to act at a distance of $\frac{D}{2}$. However, most generally, both contributions must act at the distance between their center of mass and the top of the robot's cross-section at the collapse point, which may or may not be the top of the robot given the placement of the actuators. Therefore, the moment to collapse the robot is now described by 

\begin{equation} \label{eq:comp collapse moment}
M_{collapse} = \frac{1}{2}\frac{P\pi D^2K}{4} \pm \frac{F_eK}{2} + \sum_{i=1}^{n}P_{act,i}A_{act,i}r_{act,i},
\end{equation} where $K$ is the distance from the center of the robot to the collapse point, i.e., the top-most point on the entire robot, $P_{act,i}$ is the pressure of actuator $i$, $A_{act,i}$ is the area of actuator $i$, and $r_{act,i}$ is the distance from the center of actuator $i$ to the collapse point, and there are $n$ total actuators. It should be noted that the actuators, while formed from a circular tube, can be fabricated to have non-circular cross-sectional areas. However, provided that the pressure and area of that actuator at the robot's last point of support is known, the model can still be used.

\subsubsection{Experimental Assumptions}
We will be applying this model to an unsteered robot and robots steered into curves using series pouch motor actuators (SPMs)~\cite{CoadRAM2020}. These experiments share several assumptions and commonalities outlined below.

All robots were manufactured from the same material mentioned in Section~\ref{fabrication straight}. The only difference in fabrication is that these robots had their extra flap of material at the seam removed to reduce the overall weight. When the robots were in their two layer format, the unsteered robot had a measured weight of approximately 0.02~kg/m and the SPM-steered robot had a measured weight of approximately 0.04~kg/m.

The true shape of the robot and the length of its discretized segments were determined using motion capture LEDs (PhaseSpace, San Leandro, CA). The discrete weight of the motion capture LEDs, along with distributed weights from long strips of tape and actuators, can be added into Equation~\ref{eq:current moment}. This is done for discrete weights by adding the weight multiplied by its distance from the potential collapse point along the $z$-axis. Distributed weights must be added in as the weight per unit length multiplied by the length of the distributed weight, which is assumed to be equal to the length of the segment.

For all robots tested in this section, we used the motion capture LEDs to track the position and orientation of a 3D-printed coordinate axis fixed at the base of the robot. We did this to align the motion capture data with the axes used in our equations. We placed this coordinate axis on a shelf over top of the robot, shown in Figures~\ref{straight with sensors} and \ref{actuated figure}, and later subtracted the vertical distance between the coordinate axis LEDs and the robot LEDs (0.11~m) to align the data for ease of analysis. Each LED and the 1~m-long wire it is connected to has a mass of 0.0036~kg, which is significant compared to the weights of the robots. We also assumed that the full length of an LED's cable was concentrated at the same location as the LED. Generally, the cables hung further back than their associated LED, so this is an overestimate of their moment contribution. 

Additionally, our model assumes that what we are measuring with the motion capture LEDs is the central axis of the robot. For the unsteered robot, the LEDs are located on top of the robot, parallel to the central axis. As this robot is reasonably straight, this does not result in a significant difference between the calculated length and the true length. However, the steered robots have the LEDs placed along their inner curve. This results in the robot being modeled as slightly shorter than it actually is. Given the short length of the robots (less than 1~m) and their small diameters (less than 5~cm), we assumed that this difference was minimal. Occasionally, if an LED was missing from the dataset due to not being visible to the motion capture system, we used nearby LEDs to estimate its position. 

We lengthened all robots in these experiments by pushing them off the table. The table edge acted as the last point of support, and we measured its location using the LEDs on the coordinate axis. When pushing the robots, we ensured that the segment's central axis was in line with the $z$-axis on our physical coordinate axis. We lengthened the robots until either collapse occurred or the last LED on the robot reached the edge of the table, i.e., the last point of support. In either case, the time at which the event occurred is referred to as the instant of interest. We identified this instant in the data by syncing the data collection with a video of the experiment and visually determining when it happened. For the case of collapse, we specifically attempted to identify the instant before collapse, which is right before the wrinkles at the robot base propagate all the way around its circumference. This is, however, only an approximation of the true instant of interest. 

\subsection{Analyzing Unsteered Robot Collapse with Position and Force Data}
\label{straight with sensors section}
To begin testing the comprehensive model, we first examine the behavior of an unsteered robot while measuring its position and tail tension. Measuring tail tension directly makes it possible to confirm the assumptions we have previously made about the value of tail tension at the instant of collapse. This allows us to test the model on a basic example before moving on to more complicated robots.

\subsubsection{Model}

This unsteered robot does not have actuators. Therefore, the current moment due to gravity acting on the robot can be described by

 \begin{equation} \label{eq:current moment straight}
M_{current} = \sum_{i=1}^{n}2\pi Dtd_i\rho gr_i.
\end{equation} The moment needed to collapse the robot will be now be equal to Equation~\ref{eq:straight collapse moment} with the added measured force inside the tail applied at a distance of $\frac{D}{2}$. This is described by 

\begin{equation} \label{eq:measured tail tension moment}
M_{collapse} =\frac{P\pi D^3}{8} \pm \frac{T_mD}{2},
\end{equation} where $T_m$ is the measured tail tension, which we assume acts at the center of the robot. Here, we are using $T_m$ to determine whether the value of tail tension described in Equation~\ref{eq:tail tension}, which is in turn used to derive Equation~\ref{eq:comp collapse moment} of our comprehensive collapse model, is an accurate estimate of the true value.

\subsubsection{Experimental Setup}
For position tracking, we placed eight motion capture markers at an interval of approximately 10~cm. This provided accurate shape tracking while still only requiring the use of one microdriver. This setup is shown in Figure~\ref{straight with sensors}(a). For force measurements, we used a small force sensor (Nano17, ATI, Apex, NC). We attached a custom hooked mount to the sensor, so that we could measure tension along the $z$-axis. We fabricated a robot with a 4.85~cm inflated diameter and tied its tail to one end of the force sensor. To best mimic the other experiments conducted in this paper, where the tail is sealed to bottom layer of fabric at the base, we tied the base end of the force sensor to a scrap piece of fabric and sealed that scrap piece of fabric to the inside of the robot on its bottom side. The two attachment points can be seen in Figure~\ref{straight with sensors}(b). In this way, we forcibly sealed the robot into two layers with the force sensor inside it. We ran the force sensor's cable out through a hole in the top side of the robot and used tape to make the hole airtight. We biased the sensor to read zero tension from the neutral position before pressurizing the robot to minimize sensor error. We then pressurized the robot to 3.45~kPa and pushed it slowly off the table until it collapsed. We did this process three times in total. Using the process described in Section~\ref{Fe section}, we found the value of $F_e$ for this robot to be 1.2~N. 

\subsubsection{Results}

\begin{figure}[tb]
\centering
\includegraphics[width = \columnwidth]{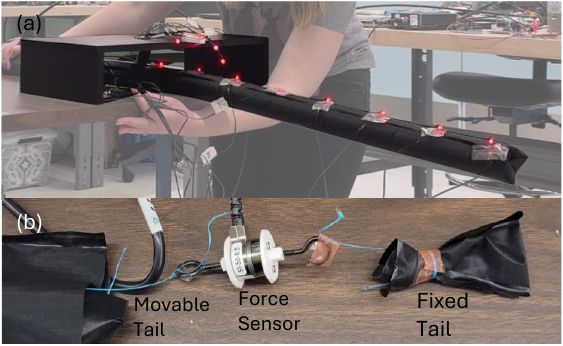}
\caption{Setup for experiment measuring tail tension and vine robot shape at the approximated instant of collapse for an unsteered robot, used to collect the data in Table~\ref{table:straight}. (a) We placed eight LEDs along the robot's length. We used three additional LEDs to track a 3D-printed coordinate axis to align the motion capture data with a reference frame fixed in the room. (b) We tied our force sensor to the robot tail and a scrap piece of fabric that was then sealed to the inside of the robot to measure the tail tension. The force sensor wire exited the side of the vine robot body through a taped-shut hole.
}
\label{straight with sensors}
\end{figure}

\begin{table}
\begin{center}
\begin{tabular}{ |c|c|c|c| }
\hline
Trial & Force (N) & Calculated Moment (Nm) & Key Metric (\%) \\
\hline
1 & 1.69 & 0.1267 & 111.4 \\
\hline
2 & 1.73 & 0.1187 & 105.3 \\
\hline
3 & 1.89 & 0.1234 & 113.3 \\
\hline
\end{tabular}
\end{center}
\caption{Measured tail tension force and calculated moment due gravity based on measured vine robot shape at instant of collapse for an unsteered robot. 
We used the measured tail tension force to determine the moment due to gravity at which we predict the robot should collapse. The calculated moment due to gravity based on the measured shape for each trial is shown as a percentage of this predicted collapse moment due to gravity for that trial. For all three trials, the calculated value is a good estimate (within 15\%) for the predicted value, indicating that our model is accurate, given full knowledge of the tail tension value.}
\label{table:straight}
\vspace{-3ex}
\end{table}

All data from the trials is presented in Table~\ref{table:straight}. Table~\ref{table:straight} shows the calculated moment based on our shape information and the key metric. The key metric defining the model's success is the calculated moment at the instant of interest based on our shape measurements divided by the moment that should cause collapse the robot based on measured tail tension. An ideal model would have a value of 100\%. For all three trials, our key metric is within 15\% of a perfect match. From this, we conclude that our method of calculating the moment acting on the robot is reasonably accurate. However, it is important to note that, even with shape and force sensing, we still achieve more than 10\% error on two of the three trials.

The primary source of error in these trials is the assumption that the weight of each LED cable is concentrated at its associated LED. In Figure~\ref{straight with sensors}, the cables toward the base of the robot hang further forward than modeled, and the cables near the top of the robot hang further back than modeled. As previously stated, this creates an overestimate of the current moment acting on the robot. This is likely why all trials found collapse to occur at a slightly higher moment than the predicted one. The other source of error is that we are taking a visual approximation of when collapse occurred, which could be slightly off from when it actually occurred. These issues are also relevant for Sections~\ref{steered section} and \ref{new demo section}. However, given that we have reasonable agreement between our results and the model, and the cables could play a significant role in the observed overshoot, we believe this method is suitable for use on our curved robots. In subsequent sections, we will continue to use 15\% as our guideline for a good prediction made by our models.  

\subsection{Steered Robot Collapse}
\label{steered section}

Next, we discuss how actuators affect the collapse length of a robot from any shape those actuators can cause the robot to form, starting with a curve made by a single actuator along the robot's full length. This allows us to test the model on a simple actuated robot and determine if we can apply it to vine robots with three actuators.

\subsubsection{Model}
\begin{figure}[tb]
\centering
\includegraphics[width = 0.75\columnwidth]{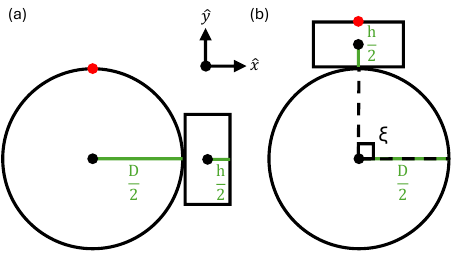}
\caption{Cross-sectional diagram of a vine robot with one series pouch motor (SPM) actuator attached to it, as in Figure~\ref{actuated figure}. (a) When the SPM is on the side of the robot, the top of the main tube (marked in red) is still the point about which collapse will occur. Moments will be taken about this point. (b) When the SPM is at the top of the robot, the top of the SPM is the new point about which collapse will occur and moments will be taken.}
\label{SPM diagram}
\vspace{-3ex}
\end{figure}

While we will continue to view the robot as a series of discretized segments, we must now account for the weight added by an actuator, as well as the stiffening effect of its pressure. The presence of actuators changes the cross section of the robot at the point of collapse, which modifies our value of $M_{collapse}$. Two examples of this new cross section for a single curvature curve can be seen in Figure~\ref{SPM diagram}. 

For these experiments, a single SPM is located either on the side of the robot, as shown in Figure~\ref{SPM diagram}(a), or on the top of the robot, as shown in Figure~\ref{SPM diagram}(b). More generally, we will say this SPM and its corresponding LED is located at a rotational angle $\xi$ about the positive $z$-axis. When the SPM is on the side of the robot, the point where collapse occurs remains the top of the robot body. However, when the SPM is on top of the robot, the top of the SPM becomes the new point about which collapse occurs. These locations are marked with red dots. These actuators are made by putting small heat seals across half the width of the inflated tube at regular intervals such that the tube inflates as a series of connected squares. When inflated, the SPM shortens, curling the robot in that direction. While SPMs have half the lay-flat diameter of the robot, the partial seals that allow them to contract force them into a flatter cross section than an inflated tube would have. Therefore, we approximate them as having rectangular cross sections, whose height and width can be measured directly to determine their area. We now have two values for $M_{collapse}$, depending on whether the actuator is located at the top or the side of the robot:

\begin{equation} \label{eq:actuated straight collapse moment side}
M_{CSide} = \frac{1}{2}\frac{P\pi D^3}{8} + \frac{P_{act}A_{act}D}{2} \pm \frac{F_eD}{4},
\end{equation} and

\begin{equation} \label{eq:actuated straight collapse moment top}
M_{CTop} = \frac{1}{2}\frac{P\pi D^2}{4}\left(\frac{D}{2}+h\right) + \frac{P_{act}A_{act}h}{2} \pm \frac{F_e}{2}\left(\frac{D}{2}+h\right),
\end{equation} where \(h\) is the measured height of the actuator at the given pressure. 

Given the robot's weight and the weight of its actuator, the current moment of the robot due to gravity is now

\begin{equation} \label{eq:current moment with actuators}
M_{current} = \sum_{i=1}^{n}3\pi Dtd_i\rho gr_i.
\end{equation} 

\subsubsection{Experimental Setup}
\begin{figure}[tb]
\centering
\includegraphics[width = \columnwidth]{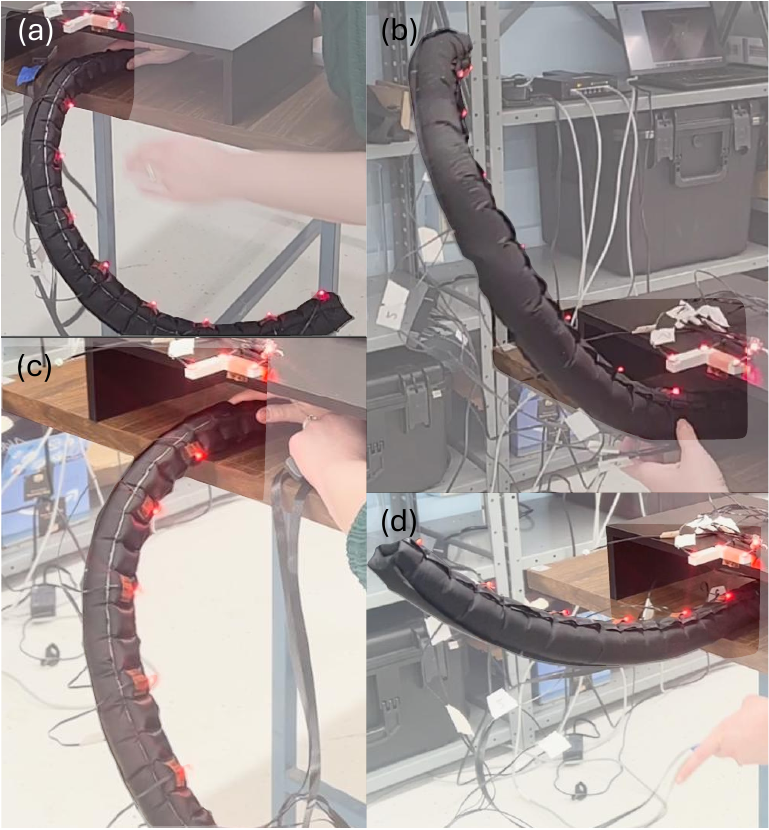}
\caption{Experimental setup for steered robot tests with a 4.04~cm diameter fabric robot pressurized to 3.45~kPa, used to collect the data in Table~\ref{table:actuated alt}. Images show the instant of interest at which collapse potential was evaluated for the robot with an actuator pressure of 17.34~kPa with $\xi$ values of (a) \mbox{$0^{\circ}$} and (b) \mbox{$90^{\circ}$}. Images also show the instant of interest for the robot with an actuator pressure of 6.34~kPa with $\xi$ values of (c) \mbox{$0^{\circ}$} and (d) \mbox{$90^{\circ}$}. In these trials, only robots with a $\xi$ of \mbox{$90^{\circ}$} collapsed. All others just deflected.}
\label{actuated figure}
\end{figure}

In these experiments, we pushed a robot with an inflated diameter of 4.04~cm and an internal pressure of 3.45~kPa off the table at two curvatures and two rotational angles, $\xi$, for a total of four configurations. We created our two curvatures by inflating the SPM to 17.24~kPa for a tighter curvature and 6.89~kPa for a looser curvature. We pushed the robot off the table at rotational angles of \mbox{0$^{\circ}$} and \mbox{90$^{\circ}$}. We ran three trials for each configuration. The robot had a single SPM with a lay-flat diameter half that of the robot. This SPM was attached with a long strip of tape weighing 0.0073~kg/m, which we accounted for in our calculation of $M_{current}$ in the manner described in Section~\ref{comprehensive model section}. We placed LEDs every three SPMs (approximately 0.1~m) along the robot and used a total of eight LEDs. The robots at their final instant of interest can be seen in Figure~\ref{actuated figure}. 

We calculated $F_e$ for the two curvatures by measuring the pressure for that robot to grow with its SPM a few centimeters from its starting length of less than 1~m. We then multiplied that pressure by the robot's cross-sectional area. We found the value of $F_e$ to be 4.5~N for the 17.24~kPa SPM configuration and 3.8~N for the 6.89~kPa SPM configuration. When the SPM was inflated to 17.24~kPa, the pouches had a height and area of approximately 1.1~cm and 2.8~cm\(^2\). When the SPM was inflated to 6.89~kPa, the pouches had a height and area of approximately 1.0~cm and 2.8~cm\(^2\). We found these values by measuring multiple pouches along the SPM and averaging their respective heights and lengths. 

\subsubsection{Results}
Here, the key metric defining the model's success is the calculated moment at the instant of interest based on our measurements divided by the moment that should cause collapse the robot based on eversion tail tension. Again, an ideal model would have a value of 100\%, but we define success as being within 15\% of that value. In our experiments, all three trials of a given configuration behaved the same, that is, they either all collapsed or all did not collapse. The two configurations where $\xi$ was \mbox{$0^{\circ}$} both resulted in significant deflection, but ultimately no collapse when the eighth LED reached the last point of support. Their key metric for these configurations was nowhere near 100\%, so collapse was not expected. 

The two configurations where $\xi$ was \mbox{$90^{\circ}$} did collapse, but they did so when they were not expected to based on the model that includes actuator pressure. Closer inspection of the video footage revealed that the robots were collapsing only when the flat seal between SPM pouches was at the last point of support, meaning, at that instant, the SPM was contributing to $M_{current}$ with its weight, but it was not contributing to $M_{collapse}$ with its pressure. If we instead look at the definition of $M_{collapse}$ from Equation~\ref{eq:tail tension moment}, the key metric becomes much closer to 100\%, making collapse an expected outcome.

These results are shown in more detail in Table~\ref{table:actuated alt}. We present the mean percentage of the key metric achieved for all three trials from each configuration for the models both without (Equation~\ref{eq:tail tension moment}) and with actuator pressure (Equations~\ref{eq:actuated straight collapse moment side} ($\xi = \mbox{$0^{\circ}$}$) and \ref{eq:actuated straight collapse moment top} ($\xi = \mbox{$90^{\circ}$}$)), the standard deviation for these means, and a note for which model accurately predicts the behavior, i.e., which model has a key metric within 15\% of 100\% only when the robot collapsed. As highlighted in the table, the two different models are necessary in order to accurately predict the behavior of all four configurations. 

\begin{table}
\begin{center}
\begin{tabular}{ |c|c|c|c| }
\hline
$\xi$, $P_{act}$ & Key Metric & Key Metric & Which Model \\
(\mbox{$^{\circ}$}, kPa) & w/o $P_{act}$ (\%) & w/$P_{act}$ (\%) & Matches Behavior \\
\hline
0, 17.24 & 41.8$\pm$8.9 & 20.2$\pm$4.3 & Both \\
\hline
90, 17.24 & 95.3$\pm$10.1 & 52.4$\pm$5.6 & w/o $P_{act}$ \\
\hline
0, 6.89 & 90.3$\pm$7.4 & 61.2$\pm$5.0 & w/$P_{act}$ \\
\hline
90, 6.89 & 96.7$\pm$15.2 & 59.3$\pm$9.4  & w/o $P_{act}$ \\
\hline
\end{tabular}
\end{center}
\caption{Collapse results for steered robots. Each configuration is defined by its rotational angle, $\xi$, and its actuator pressure, $P_{act}$. Each configuration has two maximum values for its modeled moment due to gravity required to collapse based on the eversion tail tension model, determined by whether or not the additional restoring moment due to the pressure in the actuator is included. For each configuration, we present the mean over three trials of the percentage of each maximum value that the calculated moment due to gravity based on the measured shape of the robot was at the instant of interest. Also presented is the standard deviation for that mean, and which model predicts the correct outcome to within 15\%. From this, we see that the model without $P_{act}$ is more accurate when the robot is steering up and the model with $P_{act}$ is more accurate when the robot is steering to the side.}
\label{table:actuated alt}
\vspace{-3ex}
\end{table}

While the configurations where $\xi$ was \mbox{$90^{\circ}$} always resulted in a collapse between pressurized pouches, we did not observe this effect in the configurations where $\xi$ was \mbox{$0^{\circ}$}. The heat seal provides some stiffening to the robot which, when on the robot's side, likely helps prevent wrinkle propagation to some degree. When the stiffening is on the top of the robot, it likely does not contribute much because the wrinkle has already propagated most of the way around before the stiffened area is reached. This may explain why the robot did not collapse when $\xi$ was \mbox{$0^{\circ}$} and the SPM was inflated to 6.89~kPa, despite one of the models predicting that it should have.

These experiments showed that our models can predict the collapse of curves steered by a single SPM. The difference in behavior based on where the SPM is located demonstrates the need for multiple models to accurately predict vine robot behavior. With this information, we move on to demonstrating how our models can predict the behavior of vine robots with multiple actuators. 

\section{Demonstrations}
In this section, we present two demonstrations of our model's ability to predict vine robot behavior in specific tasks. For the unsteered demonstration, we reanalyze the demonstration originally presented in \cite{mcfarland2023collapse} through the lens of the new models incorporating tail tension as described in Section~\ref{previous work}. For the steered demonstration, we perform a similar task to the unsteered demonstration using a robot steered with three SPMs and model its behavior using the method described in Section~\ref{comp collapse section overall}. In both demonstrations, the models are able to accurately predict the outcomes.

\subsection{Unsteered Demonstration} \label{old demo section}
In \cite{mcfarland2023collapse}, they presented a demonstration of a basic task vine robots must be able to do in the field: use a straight-line path to cross a gap in the environment. They showed the robot failing to complete the task with arbitrarily chosen parameters, and then showed that by individually increasing the growth angle, the diameter, and the pressure, they could achieve the task. Here we analyze those results with our updated model. 

\begin{figure*}[tb]
\centerline{\includegraphics[width = \textwidth]{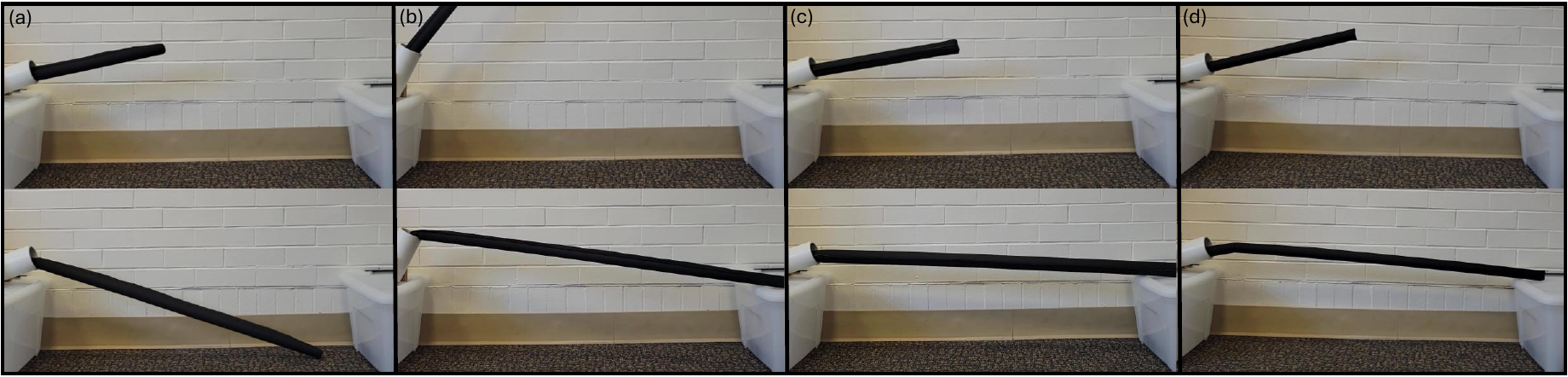}}
\caption{Demonstration of our collapse model predicting the performance of an unsteered vine robot as it attempts to grow over a 0.95~m gap (modified from \cite{mcfarland2023collapse}), corresponding to modeled results shown in Table~\ref{table:olddemo}. (a) With an initial set of parameters, the vine robot collapses when trying to cross the gap. However, increasing the (b) growth angle, (c) diameter, or (d) internal body pressure results in the robot succeeding in the task. The robots generally collapsed at a shorter length than anticipated using the original model from~\cite{mcfarland2023collapse} without tail tension. Using our model with tail tension, it is clear why the parameters had to be increased more than expected in order for the robot to succeed.}
\label{olddemo}
\vspace{-3ex}
\end{figure*}

The authors of \cite{mcfarland2023collapse} created a setup with two boxes of equal height set 0.95~m apart and placed their robot launcher at the edge of the leftmost box. The robot was grown through the launcher until it collapsed or reached the target, that is, the right box. They controlled the growth speed to keep it quasistatic by manually pinching the two layers of material together behind the launcher. In Table~\ref{table:olddemo}, we present the parameters for each robot, along with the collapse length predicted by their original model, which had no tail tension, as well as our model for collapse when the robot experiences eversion tail tension. The intentional collapse configuration, shown in Figure~\ref{olddemo}(a), had a growth angle of \mbox{20$^{\circ}$}, a diameter of 3.24~cm, and a pressure of 4.14~kPa, and it was modeled to collapse at 0.82~m. It did indeed collapse at some length shorter than 0.95~m and failed the task. When we analyze this value with tail tension, we see that this robot should have failed according to either of the models. 

The authors of \cite{mcfarland2023collapse} then moved on to robots that could succeed in the task. They began by changing the angle to \mbox{65$^{\circ}$}. This robot grew to a sufficiently long length such that when it collapsed it fell onto the target, which can be seen in Figure~\ref{olddemo}. However, the original model showed that \mbox{55$^{\circ}$} should have also been sufficiently long, but the robot failed at that angle. However, when we analyze this trial with tail tension, we can see why that occurred. The collapse length for the robot growing at \mbox{65$^{\circ}$} with eversion tail tension is sufficiently long that it should have succeeded, but this is not true for that robot growing at \mbox{55$^{\circ}$}.

\begin{table}
\begin{center}
\begin{tabular}{ |c|c|c|c| }
\hline
$\gamma$, $D$, $P$ & No $T_t$ & Eversion $T_t$ & Which Model \\
(\mbox{$^{\circ}$}, cm, kPa) & (m) & (m) & Matches Behavior \\
\hline
20, 3.24, 4.14  & 0.82 & 0.69 & Both \\
\hline
65, 3.24. 4.14 & 1.20 & 1.01 & Both \\
\hline
20, 4.04, 4.14 & 1.05 & 0.83 & Both \\
\hline
20, 3.24, 10.34 & 1.31 & 1.00 & Both \\
\hline
\end{tabular}
\end{center}
\caption{Modeled collapse length for unsteered demonstration robots with various models. The robot attempted to cross a 0.95~m gap. In the first configuration, the robot collapsed at a length too short to reach across the gap, while in the other three configurations, it collapsed at a length just long enough to reach across the gap. For each configuration, we present the predicted collapse length for our model without tail tension, with eversion tail tension, and which model matches the behavior. While the eversion tail tension model does not technically predict a long enough collapse length to succeed in the task for the third configuration, it is within 15\% of the length needed to succeed. Given that the robot did succeed, we consider the model to have correctly determined the outcome.}
\label{table:olddemo}
\vspace{-3ex}
\end{table}

The authors of \cite{mcfarland2023collapse} then adjusted the angle back to \mbox{20$^{\circ}$} and changed to a robot with a diameter of 4.04~cm. This robot succeeded as expected, which is shown in Figure~\ref{olddemo}. When we analyze this robot with our new models, we see that the collapse length for eversion tail tension is 0.12~m short of what the robot would have needed to succeed. However, this is within 15\% of the needed value to cross the gap, so the model did predict the outcome correctly. 

Finally, the authors of \cite{mcfarland2023collapse} changed back to their original 3.24~cm robot with a growth angle of \mbox{20$^{\circ}$} and adjusted the pressure up to 10.34~kPa. This robot also succeeded in the task, which is shown in Figure~\ref{olddemo}. However, once again this was a greater increase than they had anticipated needing to make for the robot to succeed. It barely grew to a sufficiently long length for the task, even though the model predicted it would have a collapse length of 1.31~m. However, when we evaluate this situation with tail tension we see why. If the robot was experiencing eversion tail tension, it would be able to succeed in the task, but any higher amount of tension could lead to failure. Since they note that they were controlling the growth speed by pinching the tail, this could have made it more challenging to achieve a low enough tail tension for the robot to succeed.

Looking at this unsteered demonstration, it is clear that tail tension had an influence on the results. Although deflection may still have played a role as was explained in previous experiments, this demonstration highlighted the need for a more advanced model that incorporates shape information, especially because few tasks are as simple as this one was.

\subsection{Steered Demonstration} \label{new demo section}
Finally, we applied our model to a vine robot that can steer in any direction, since this is the design that would most likely be used in the field and would most benefit from the use of our model. Here, we designed a task similar to the one in Section~\ref{old demo section}. The robot had to grow over a gap 0.59~m wide. The far side was a cart slightly taller than the starting point, meaning the robot had to steer to succeed in the task. To maximize the curve created by the actuators, only the top two actuators were inflated while the third was left uninflated.

Since two actuators were inflated in this trial, we had to adjust our equation for $M_{Collapse}$ to account for the contribution from both actuators at their respective positions. The actuators were arranged such that the uninflated one was on the bottom and the other two were \mbox{120$^{\circ}$} apart from it in each direction around the robot's circumference. This arrangement was previously shown in Figure~\ref{inflated tube diagram}, though it should be noted the modeled shape of the actuators is more similar to the way they are shown in Figure~\ref{SPM diagram}. The collapse point was modeled as the top of the robot. This new collapse moment accounting for the actuator contributions came out to be

\begin{equation} \label{eq:demo collapse moment}
M_{Collapse} = \frac{1}{2}\frac{P\pi D^3}{8} + 2P_{act}A_{act}\left(\frac{D}{4}-\frac{h}{4}\right) \pm \frac{F_eD}{4}.
\end{equation}

\begin{figure}[tb]
\centering
\includegraphics[width = \columnwidth]{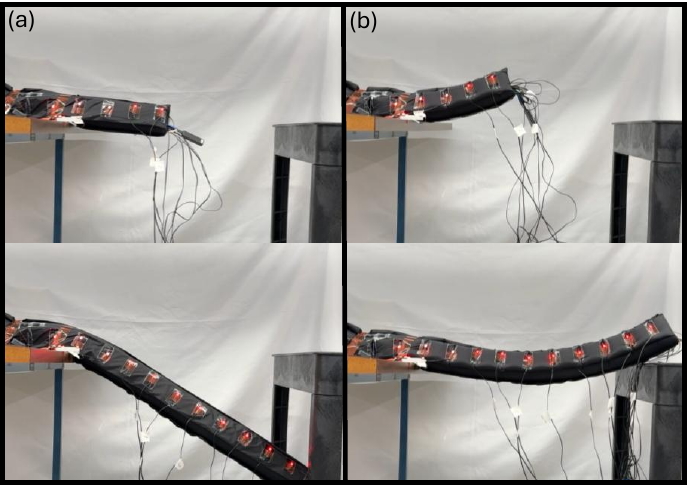}
\caption{Demonstration of our collapse model predicting the performance of a three-actuator steered vine robot as it attempts to grow over a 0.59~m gap, corresponding to modeled results shown in Table~\ref{table:newdemo}. (a) When the actuators are at 0~kPa, the vine robot deflects downward and collapses before completing the task. (b) When the two top actuators are inflated to 3.45~kPa, the robot curves slightly upward, and it then deflects down onto the other side of the gap.}
\label{New Demo}
\end{figure}

For this demonstration, our vine robot had an inflated body diameter of 8.1~cm with three actuators being half that in diameter. For all trials, the robot had an internal body pressure of 6.89~kPa. We set the top two actuators to 0~kPa (for the first configuration), shown in Figure\ref{New Demo}(a), and 3.45~kPa (for the second configuration), shown in Figure~\ref{New Demo}(b). 

For all trials, we once again accounted for the weight of the LEDs, the tape attaching the actuators to the robot, and the two pieces of tape covering up the SPM seams as they were sewn over for extra strength. We deemed the weight of the thread negligible. It should also be noted that, since this robot was allowed to grow, the robot was supporting the weight of the 0.016~kg microdrivers that power the LEDs throughout growth. This likely increased deflection. At the instant of collapse, one microdriver was supported by a block placed in the environment for that purpose, while one hung freely at the edge of the cart due to the way the wires moved around the cart as the robot deflected. Therefore, the weight of one microdriver was added in at a distance of 0.59~m.

We analyzed collapse looking at the value of $M_{collapse}$ described in Equation~\ref{eq:demo collapse moment} (with $P_{act}$) and Equation~\ref{eq:tail tension moment} (without $P_{act}$. From the way the SPMs were attached, one of the top two actuators had a seam at the collapse point while the other had an inflated pouch near the collapse point, so some support from the SPM could be possible. However, we know from previous trials that the robot is more likely to collapse between the pouches. For these trials, we found the value of $F_e$ to be 9.4~N and 14.1~N corresponding to when the top two SPMs were at 0~kPa and 3.45~kPa, respectively. Additionally, the measured height and area for the 3.45~kPa SPMs were 2~cm and 10~$cm^2$. 

\begin{table}
\begin{center}
\begin{tabular}{ |c|c|c|c| }
\hline
$P_{act}$ & Key Metric & Key Metric & Which Model \\
(kPa) & w/o $P_{act}$ (\%) & w/$P_{act}$ (\%) & Matches Behavior \\
\hline
0 & 85.4 & 85.4 & Both \\
\hline
3.45 & 55.3 & 50.1 & Both \\
\hline
\end{tabular}
\end{center}
\caption{Modeled collapse length for steered demonstration robots with various models when applicable. The robot attempted to cross a 0.59~m gap with and without modeling the restorative force of the two inflated actuators. For both configurations, we present the same key metric from Table~\ref{table:actuated alt} when using the model without actuator pressure, and when using the model with actuator pressure when applicable, as well as which models describe the result. In both configurations, the actuator contribution is so small that both models accurately predict the outcomes. Therefore, either could be used.}
\label{table:newdemo}
\vspace{-3ex}
\end{table}

The results are shown in Table~\ref{table:newdemo}. When the SPMs were not inflated, the robot's weight caused it to start deflecting significantly off course, even in a task with such a short distance. This can be seen in Figure~\ref{New Demo}(a). As the robot deflected, it grew longer until it finally collapsed. While it did collapse at a length long enough to cross the gap, it had already deflected too far off course to succeed. At the instant of collapse, the robot had a moment measuring 85\% of the eversion tail tension collapse moment. This result would lead us to expect collapse. From this trial, we showed that the model can predict collapse fairly accurately for this straight robot carrying many additional weights. However, in real time, a user would observe task failure long before the occurrence of collapse. 

When the SPMs were inflated to 3.45~kPa, the robot curved slightly upward, which can be seen in Figure~\ref{New Demo}(b). It did not experience collapse and ultimately deflected down onto the cart. Accounting for the actuators, the moment acting on the robot was 50\% of the eversion tail tension moment. From this, we would predict the robot was nowhere near collapse. Without accounting for the actuators, the calculated moment acting on the robot was 55\% of the eversion tail tension moment. By either model, we would have expected the robot to succeed in the task assuming it did not deflect off course. From this demonstration, we see that the model can be applied to a vine robot design that would be used in field applications.    

\section{Discussion}
In this paper, we analyze trends in vine robot collapse behavior, present a model that predicts collapse given the robot's true shape information, and show how this work can be applied to robots completing a real-world task.
Even with tail tension applied to the model, the original recommendations for how to prevent vine robot collapse from \cite{mcfarland2023collapse} still hold: A vine robot will collapse at a longer length if it is grown at a higher growth angle, operated at a higher pressure, or fabricated with a larger diameter. Our predictions for collapse length also hold when we inflate additional supports along the robot. However, while pressure can be changed during operation, diameter will remain fixed. Growth angle will be largely determined by the environment for an unsteered robot. These trends can help with design, but ultimately cannot heavily influence the behavior of a robot once it is deployed.

The comprehensive collapse model would be best deployed in combination with real-time vine robot shape sensing. This would allow a user to monitor the increasing moment due to gravity acting on the robot. If it exceeds the moment large enough to cause collapse based on the value of eversion tail tension, the user could then redirect the robot, increase its pressure, or otherwise modify its parameters to lessen the chances of failure before the task is complete. If the approximate path was known in advance, the model could also be used to design the optimal robot that would complete the desired task. Additionally, by installing a force sensor inside the vine robot base, a user could use real-time tail tension measurements instead of needing to estimate $F_e$ for different configurations. 

\section{Conclusion}
In this paper, we present an extended analysis of the collapse model for unsteered vine robots from \cite{mcfarland2023collapse}, and we also present our comprehensive collapse model. We apply this more advanced model to unsteered vine robots with inflated tube supports, unsteered vine robots whose shape we measure directly, and SPM-actuated steering vine robots. We show that adding tail tension to our model greatly improves its accuracy for unsteered robots, in addition to being accurate for steered robots. We analyze how this improves our understanding of the demonstration from \cite{mcfarland2023collapse}, and we demonstrate how our model can be used to understand the behavior of a vine robot with three SPMs completing a task.

From this work, we can conclude that if we know the shape of the robot, we can form some basic guidelines of the path it should grow along to be successful in a task. Most current field-use vine robots do not retract, which means repeating a path would require the user to pull the robot out and grow again. With this model, users could have greater confidence that the robot is growing in such a way as to complete the task on the first try. This model could also be useful for other thin-walled inflatable robots.

Other factors we wish to consider are the dynamic impacts of growth, as well as the effect of the material selection and manufacturing decisions. We are also interested in modeling the movement of the tail inside the robot and its effect on collapse and curvature. This effect will likely be more prominent as the robot achieves higher curvature and also multi-curvature shapes. With all this information, we aim to form a path planning algorithm from this model to create a more intuitive and useful user experience for operating the vine robot.


%



\ifCLASSOPTIONcaptionsoff
  \newpage
\fi



%

\bibliographystyle{IEEEtran}
\bibliography{library}



%







\end{document}